\documentclass{article}
\usepackage{spconf,amsmath,graphicx}
\usepackage{xcolor}
\usepackage{xfrac}
\usepackage{booktabs,multirow}
\usepackage{gensymb}
\usepackage{tikz}

\newcommand{\minisection}[1]{\vspace{0.04in} \noindent {\bf #1}\ \ }

\title{Learning Illuminant Estimation from Object Recognition}
\name{Marco Buzzelli $^a$\sthanks{Corresponding author: marco.buzzelli@disco.unimib.it}, Joost van de Weijer $^b$, Raimondo Schettini $^a$}
\address{
\begin{tabular}{@{}c@{}}
  $^a$ Dipartimento di Informatica, Sistemistica e Comunicazione\\
  Universit\`{a} degli Studi di Milano-Bicocca\\
  Viale Sarca 336, Milan 20126, Italy
  \relax
\end{tabular}
\hskip 2em
\begin{tabular}{@{}c@{}}
  $^b$ Computer Vision Center\\
  Universitat Aut\`{o}noma de Barcelona\\
  Bellaterra, Spain
  \relax
\end{tabular}
}
\begin{document}
%
\maketitle
%
\begin{abstract}
In this paper we present a deep learning method to estimate the illuminant of an image. Our model is not trained with illuminant annotations, but with the objective of improving performance on an auxiliary task such as object recognition.
To the best of our knowledge, this is the first example of a deep learning architecture for illuminant estimation that is trained without ground truth illuminants.
We evaluate our solution on standard datasets for color constancy, and compare it with state of the art methods. Our proposal is shown to outperform most deep learning methods in a cross-dataset evaluation setup, and to present competitive results in a comparison with parametric solutions.
\end{abstract}
%

\section{Introduction}
\label{sec:introduction}
\begin{tikzpicture}[remember picture,overlay]
\node[anchor=south,yshift=15pt] at (current page.south) {\fbox{\parbox{\dimexpr\textwidth-\fboxsep-\fboxrule\relax}{
\scriptsize \textcopyright \,
Copyright 2018 IEEE. Published in the IEEE 2018 International Conference on Image Processing (ICIP 2018), scheduled for 7-10 October 2018 in Athens, Greece. Personal use of this material is permitted. However, permission to reprint/republish this material for advertising or promotional purposes or for creating new collective works for resale or redistribution to servers or lists, or to reuse any copyrighted component of this work in other works, must be obtained from the IEEE. Contact: Manager, Copyrights and Permissions / IEEE Service Center / 445 Hoes Lane / P.O. Box 1331 / Piscataway, NJ 08855-1331, USA. Telephone: + Intl. 908-562-3966.
}}};
\end{tikzpicture}
Color constancy is the ability of human beings to recognize the colors of objects independently of the characteristics of the light source. Computational color constancy aims to first estimate the illuminant and subsequently use this information to correct the image, to display how it would appear under a canonical illuminant~\cite{forsyth1990novel}. The class of ``learned'' methods is among the most successful illumination estimation methods to date~\cite{gijsenij2011computational,akbarinia2017colour}, and typically relies on a training set of images which are labeled with the respective scene illuminant.
Although the human visual system is often compared to a machine learning algorithm, during evolution it was never presented with ground truth illuminants. Instead it is hypothesized that the ability of color constancy arose because it helped other crucial tasks, such as recognizing fruits, objects, and animals independently of the scene illuminant \cite{bramao2012contribution}.

\begin{figure}
\begin{center}
\includegraphics[width=1\linewidth]{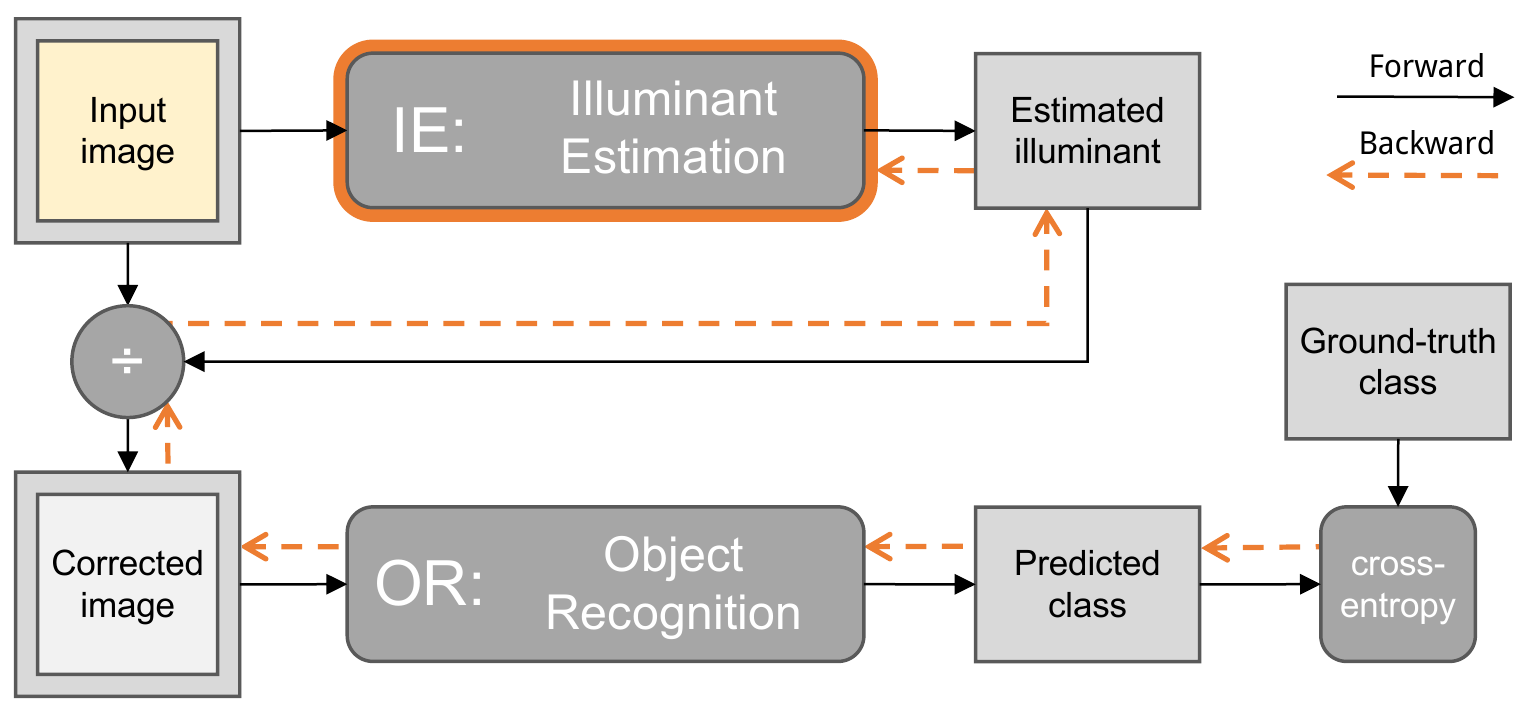}
\end{center}
   \caption{Schematic representation of our proposal for computational color constancy. The Illuminant Estimation module is trained with the objective of optimizing the Object Recognition module. At inference time, only the Illuminant Estimation module is needed.}
\label{fig:base_idea}
\end{figure}

This observation motivates us to investigate to what extent we can learn illuminant estimation as a byproduct of an auxiliary task, for which we consider object recognition.
Previous works highlighted the usefulness of object recognition in applications related to color theory, such as assessment of chromatic adaptation transforms through memory color matching \cite{smet2017study}, performing color constancy with the assistance of faces \cite{bianco2012color} or via pre-training on generic object recognition before fine-tuning for illuminant estimation \cite{lou2015color}. In our proposal we describe a training process which uses only the labels for the auxiliary task but no illuminant ground truth whatsoever.
During training, the input image will first pass through an illuminant estimation network, which estimates the scene illuminant and corrects the image accordingly. The white-balanced image will then be processed by an object classification network, which produces an estimation of the classes that are present in the image. By training both networks in an end-to-end fashion, we can effectively train the illuminant estimation network without any illuminant ground truth.
This illuminant estimation network can then be independently applied to other datasets, such as standard color constancy benchmarks, without using the object recognition network.
To the best of our knowledge this is the first learned method for illuminant estimation which does not require illuminant annotations.

Approaches to color constancy can be divided into parametric and learned.
Parametric solutions are typically based on handcrafted features which depend on few manually-tunable parameters, the most effective to date being Akbarinia et al. \cite{akbarinia2017colour} and Cheng et al. \cite{cheng2014illuminant}.
For a complete review of parametric methods for illuminant estimation we refer to~\cite{gijsenij2011computational}.
Learning-based solutions recently proved to be particularly effective when trained in a cross-validation setup:
Bianco et al. \cite{bianco2015color} developed a patch-wise illuminant estimation neural network and generate a global estimation through consensus.
Lou et al. \cite{lou2015color} propose the fine-tuning of a network that was pre-trained for both object recognition, and illuminant regression based on the predictions of other color constancy methods.
Oh et al. \cite{oh2017approaching} reformulate the problem of illuminant estimation as a classification task by clustering the target illuminants.
All these solutions require an illuminant ground truth, and are designed to be trained on the same type of images that were seen during the test phase.

\section{Proposed Method for Color Constancy}
\label{sec:proposed}
In this section, we present our learning approach to estimate the scene illuminant in the absence of any illumination ground truth data, but with label information for an auxiliary task. Here we consider object recognition as the auxiliary task, but other objectives such as object detection or semantic segmentation could be used as well.
We propose an Illuminant Estimation / Object Recognition network (IEOR), which is composed of two parts: an Illuminant Estimation module (IE) and an Object Recognition module (OR), as shown in Fig. \ref{fig:base_idea}.
Following this structure, IE is learning to predict, for any given image, a color correction that would improve OR.

\subsection{Object Recognition network}
\label{sec:or}
The Object Recognition network performs the auxiliary supervised task described in our approach, which is used to indirectly train the Illuminant Estimation network. OR takes an input RGB image, and produces a prediction of class presence. For the purpose of this paper, we will focus on vegetable recognition \cite{hou2017vegfru}, a task for which color is important as highlighted in Fig. \ref{fig:veg_examples}. Other problems for which a correct white balancing is expected to improve results might be recognition of painting styles \cite{bianco2017deeppainting}, or assessment of image aesthetics \cite{bianco2016predicting}.

In order to obtain the desired separation of the network's internal logic into IE and OR modules, with the intermediate representation being the estimated illuminant, it is necessary to make OR overly-sensitive to color variations. This would then effectively drive IE to work as a useful preprocessing step, i.e. to learn to perform the necessary color correction.
To fulfill this requirement, the following strategy is adopted:
\begin{enumerate}
\itemsep0em
\item We pre-train OR alone on the chosen auxiliary task.\\No color jittering is applied in this phase.
\item We connect IE and train the whole system end-to-end with color jittering as described in Sec. \ref{sec:ie}.\\During this second phase the gradients flow through OR, but we update only the weights in IE.
\end{enumerate}
If we do not constrain the training process as explained, it might lead to unusable solutions, such as always producing a neutral illuminant, or spreading the processing without a clear distinction of the intermediate ``estimated illuminant''.
%
For the current implementation of our method, we adopt an architecture based on AlexNet \cite{krizhevsky2012imagenet}. The network weights are initialized on the task of object classification from ImageNet \cite{krizhevsky2012imagenet}, with the final fully-connected layer re-instantiated in order to match the different cardinality and task. Before feeding the image to the network, we subtract 0.5 as an approximation of the mean-image subtraction technique that allows to speed up training and fine-tuning of the neural model.

\subsection{Illuminant Estimation network}
\label{sec:ie}
The Illuminant Estimation network takes an input RGB image and predicts the scene illuminant.
For the purpose of this paper, we choose to limit the illuminant model to a diagonal matrix, i.e. $\rho^e = \left( \rho^e_R, \rho^e_G, \rho^e_B \right)^{\rm{T}}$, although alternative combinations are possible \cite{bianco2017artistic}:
\begin{equation}
\begin{bmatrix}
    R_{\text{out}}\\
    G_{\text{out}}\\
    B_{\text{out}}
\end{bmatrix}
=
\begin{bmatrix}
    1/\rho^e_R & 0         & 0 \\
    0         & 1/\rho^e_G & 0 \\
    0         & 0         & 1/\rho^e_B
\end{bmatrix}
\cdot
\begin{bmatrix}
    R_{\text{in}}\\
    G_{\text{in}}\\
    B_{\text{in}}
\end{bmatrix}
\end{equation}
This representation also matches the annotation associated to most datasets for color constancy, which provide ground truth illuminants in terms of triplets \cite{shi2000re,ciurea2003large,cheng2014illuminant}.

During the end-to-end training of the whole IEOR network, we artificially augment the input data by applying a random illuminant extracted from a Gaussian distribution with mean 1 and standard deviation $0.3$.
Once training is completed as described in Sec. \ref{sec:or}, IE can be used as a standalone network for color correction on any image, which is not necessarily depicting the classes seen during the training phase.

The IE module is also based on AlexNet, which was shown in the past to perform well on color constancy \cite{lou2015color}, and images are preprocessed by subtracting 0.5 as with the OR module.
The color-corrected image obtained by applying the estimated illuminant to the input image is clipped between 0 and 1.
This operation is necessary to avoid feeding to the pre-trained OR values in a range that was never encountered during training, and is also recreating what would happen by displaying the color-corrected image.

\section{Experiments}
\label{sec:experiments}
In these experiments we set out to evaluate how our illuminant estimation network, which does not require any illuminant ground truth, compares to other illuminant estimation methods on standard datasets. We consider two possible normalization techniques (global and channel-wise), and compare the results with state of the art methods.

\subsection{Experimental setup}
\begin{figure}
\begin{center}
\includegraphics[width=1\linewidth]{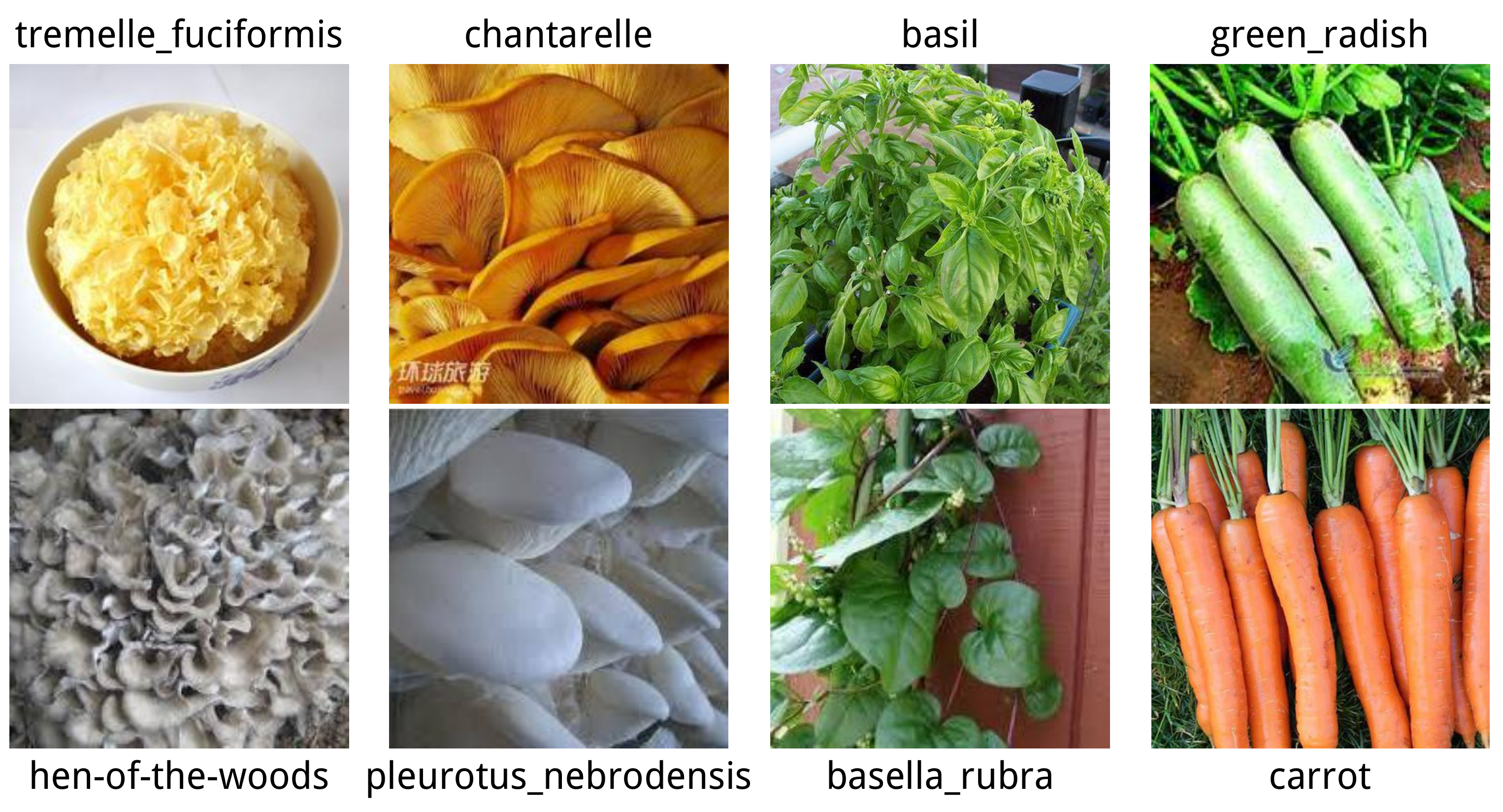}
\end{center}
   \caption{Samples from VegFru dataset, showing the importance of color in vegetable class discrimination}
\label{fig:veg_examples}
\end{figure}

The training phase requires a classification task where color is a discriminating feature, in order to properly drive the learning process. To this extent, the vegetables subset from the \textbf{VegFru} dataset \cite{hou2017vegfru} fits the desired criteria, as shown in Fig. \ref{fig:veg_examples}. It contains more than 90000 images belonging to 200 vegetable classes.
At test time we only apply the IE module (the branch related to object recognition is not used) and we evaluate our solution on two widely-adopted benchmarks for color constancy. 
\textbf{Shi-Gehler} \cite{shi2000re} is a linear reprocessing of the original RAW images from the Gehler dataset \cite{gehler2008bayesian}. It contains 568 images of different scenarios with a ColorChecker in each picture as a support for the ground truth extraction. At test time, the area corresponding to the ColorChecker is edited out to prevent the illuminant estimation algorithm from directly using it.
Results on the Shi-Gehler datasets are computed with the ground truth from \cite{bianco2015color}.
\textbf{NUS} from National University of Singapore \cite{cheng2014illuminant} contains a total of 1853 images coming from 9 digital cameras of various brands. The type of content is similar to Shi-Gehler, with both indoor and outdoor scenes, and ground truth extracted from a ColorChecker. A unique characteristic of this dataset is the presence of the same scenes taken from different cameras.
Results on NUS are reported as the average performance on each camera.

For evaluation we use the well-established angular recovery error \cite{gijsenij2011computational,cheng2014illuminant}, which compares the estimated ($\rho^e$) and reference ($\rho^r$) illuminant triplets regardless of their magnitude:
\begin{equation}
err_{ang} = \text{arccos} \left( \frac{\rho^{e\rm{T}}\,\rho^r}{||\rho^e||\,||\rho^r||} \right)
\end{equation}

\subsection{Input normalization}
In order to account for the potential discrepancies between training and test images, we adopt several techniques for input normalization.
We apply gamma correction whenever a linear dataset is used at test time, since our training classification datasets are already processed for gamma. The estimated illuminant is then re-corrected before the final evaluation with the provided ground truth.
We then devise two alternative ways to ensure the range of image values is stable, based on the generation of a support diagonal illuminant:
\begin{enumerate}
\itemsep0em 
\item Global normalization, which brings the overall average of the input image to a fixed value without changing the relationship between single channels.
\item Channel normalization, that modifies each channel independently, thus affecting the original illuminant.
\end{enumerate}

\subsection{Results}

\begin{table*}																						
\centering																						
\caption{Performance of different algorithms for illuminant estimation. Our solution can be directly compared with cross-dataset learned methods. Values reported for parametric solutions \cite{cheng2014illuminant} and \cite{funt2012removing} are selected with the best configuration of test parameters. Learned methods with in-dataset training are also reported, although a direct comparison cannot be performed.}																						
\label{tab:comparison}																						
\begin{tabular}{ll|rrrr|rrrr}																						
\toprule																						
	&	\multirow{2}{*}{\shortstack[l]{Method}}		&	\multicolumn{4}{c}{Angular error (Shi-Gehler \cite{shi2000re})}							&	\multicolumn{4}{|c}{Angular error (NUS \cite{cheng2014illuminant})}							\\		
	&			&	Mean	&	Median	&	Std	&	Max	&	Mean	&	Median	&	Std	&	Max	\\		
\midrule																						
\multirow{3}{*}{\shortstack[l]{Baselines}}	&	Unchanged		&	13.62\degree	&	13.55\degree	&	2.85\degree	&	27.37\degree	&	19.50\degree	&	18.82\degree	&	1.90\degree	&	25.83\degree	\\		
	&	Greyworld		&	7.35\degree	&	6.70\degree	&	3.78\degree	&	25.84\degree	&	4.59\degree	&	3.64\degree	&	3.57\degree	&	22.61\degree	\\		
	&	Regression		&	5.96\degree	&	5.31\degree	&	3.47\degree	&	19.88\degree	&	5.19\degree	&	3.90\degree	&	4.16\degree	&	22.07\degree	\\	
\midrule																						
\multirow{3}{*}{\shortstack[l]{Parametric}}	&	Akbarinia et al.	\cite{akbarinia2017colour}	&	3.8~~\degree	&	2.4~~\degree	&	-	&	-	&	-	&	-	&	-	&	-	\\	
	&	Cheng et al.	\cite{cheng2014illuminant}	&	3.52\degree	&	2.14\degree	&	-	&	28.35\degree	&	3.02\degree	&	2.12\degree	&	-	&	17.24\degree	\\	
	&	Funt et al.	\cite{funt2012removing}	&	3.2~~\degree	&	2.3~~\degree	&	-	&	21.7~~\degree	&	-	&	-	&	-	&	-	\\	
\midrule																						
\multirow{5}{*}{\shortstack[l]{Learned\\(cross-dataset)}}	&	\textbf{Ours (global norm.)}		&	4.84\degree	&	4.12\degree	&	3.22\degree	&	20.80\degree	&	4.88\degree	&	4.17\degree	&	3.11\degree	&	18.70\degree	\\	
	&	\textbf{Ours (channel norm.)}		&	5.48\degree	&	4.81\degree	&	3.21\degree	&	19.88\degree	&	4.32\degree	&	3.37\degree	&	3.56\degree	&	22.36\degree	\\	
	&	Joze et al.	\cite{joze2014exemplar}	&	6.5~~\degree	&	5.1~~\degree	&	-	&	-	&	-	&	-	&	-	&	-	\\	
	&	Gao et al.	\cite{gao2015color}	&	5.03\degree	&	3.39\degree	&	-	&	-	&	-	&	-	&	-	&	-	\\	
	&	Lou et al.	\cite{lou2015color}	&	4.7~~\degree	&	3.3~~\degree	&	5.3~~\degree	&	-	&	-	&	-	&	-	&	-	\\	
\midrule																						
\midrule																						
\multirow{3}{*}{\shortstack[l]{Learned\\(in-dataset)}}	&	Chakrabarti et al.	\cite{chakrabarti2012color}	&	3.59\degree	&	2.96\degree	&	-	&	21.58\degree	&	3.04\degree	&	2.40\degree	&	-	&	15.38\degree	\\	
	&	Bianco et al.	\cite{bianco2015color}	&	2.63\degree	&	1.98\degree	&	-	&	14.77\degree	&	-	&	-	&	-	&	-	\\	
	&	Oh et al.	\cite{oh2017approaching}	&	2.16\degree	&	1.47\degree	&	-	&	-	&	2.41\degree	&	2.15\degree	&	-	&	-	\\	
\bottomrule																						
\end{tabular}																						
\end{table*}																						

Table \ref{tab:comparison} reports the results of the two variants of our method on Shi-Gehler and NUS datasets, compared to different baselines and different algorithms from the state of the art.

Global normalization brings to very similar results on the two datasets, providing a robust performance assessment of our solution.
Channel normalization leads instead to contrasting conclusions on each dataset, suggesting a more challenging nature of NUS over Shi-Gehler.
As channel normalization destroys the relationship between channels, in fact, it essentially discards any potentially misleading information about the original illuminant. 
In the case of NUS, our model benefits from this preprocessing, as it is forced to estimate the proper illuminant directly from the image content.

In order to assess the effective advantage that comes from using a classification-based loss, we train the Illuminant Estimation module for direct regression (third row) over the color-augmented examples of VegFru, i.e. we train without OR. Performance on both sets shows that our classification-based strategy can lead up to a 19\% improvement in angular error, thus highlighting the relevance of the proposed approach.

The ``Learned (cross-dataset)'' block includes data-driven models that are trained on one dataset and tested on a different one, instead of adopting a more traditional training-test split of the same dataset.
Our solution was trained on the VegFru dataset, and is therefore best compared with methods from this category, which are either trained on the GrayBall dataset \cite{joze2014exemplar,lou2015color} or on SFU-Lab \cite{gao2015color}.
Our IEOR network outperforms the solutions by Joze et al. \cite{joze2014exemplar} and Gao et al. \cite{gao2015color}, while producing comparable results to Lou et al. \cite{lou2015color}.
For completeness, we also include other state of the art methods that are trained on the training-set portion of the same dataset used for evaluation (``Learned (in-dataset)''), although direct comparison with these solutions is not applicable. Parametric methods \cite{cheng2014illuminant} and \cite{funt2012removing} present different results with varying parameters and configurations. For such solutions we only report the best configuration, which was directly selected from the test set performance.

\begin{figure}
\begin{center}
\includegraphics[width=1\linewidth]{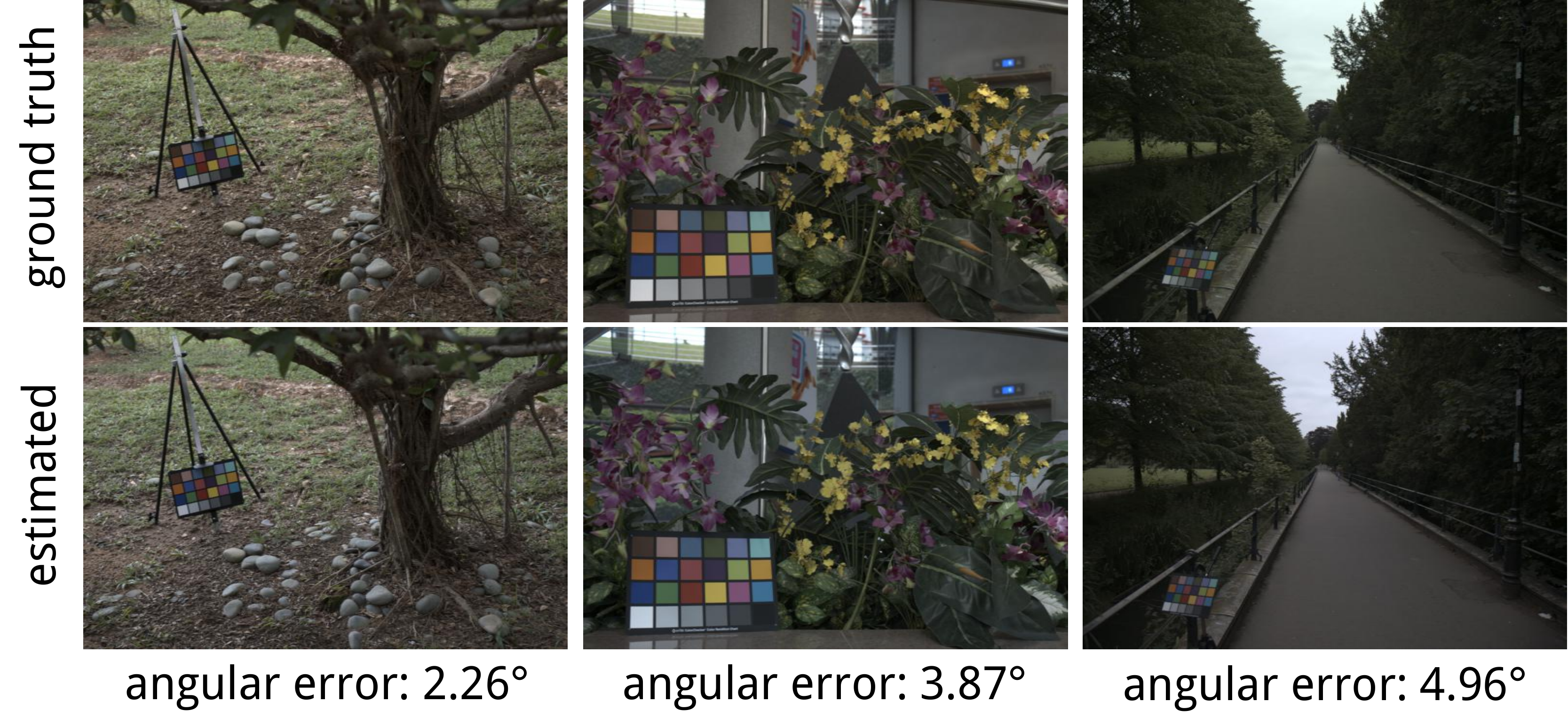}
\end{center}
   \caption{Example color corrections of our method at different levels of angular error}
\label{fig:correction_examples}
\end{figure}

Figure \ref{fig:correction_examples} shows some examples of our predictions with the corresponding angular error, in order to provide a visual guide for the interpretation of the reported performance values.

\subsection{Analysis on the training-set color bias}

The illuminant distribution present in the classification training dataset plays an important role for our proposed method.

IE implicitly learns to replicate the average illuminant of the training set. If this average is not neutral, the predicted illuminants on the test set will not, in general, match the provided ground truth.
If we use an oracle to shift our predictions to the average ground truth illuminant, we can expect to further reduce the angular error on Shi-Gehler from $4.84\degree$ to $4.60\degree$ (5\%), and from $4.32\degree$ to $4.23\degree$ on NUS (2\%).

Secondly, a high variability in the illuminant during the initial pre-training of OR may make it excessively robust to color variations, thus compromising the consequent training of IE. 
This condition is particularly hard to avoid: since dataset heterogeneity typically results in better classification performance \cite{bianco2017deeplogo}, dataset curators tend to implicitly or explicitly encourage a certain degree of illuminant variability.
To this extent, in the future we might consider preprocessing the classification-oriented training set itself for unsupervised white balancing, reducing the intra-class variability with the help of methods for saliency estimation \cite{bianco2018multiscale}.

\section{Conclusions}
\label{sec:conclusions}
We have presented a deep learning method for illuminant estimation that does not require any ground truth illuminant for training.
In order to fulfill this objective, we indirectly train our neural model with the task of improving the performance of an auxiliary object recognition network. Our strategy allows the use of powerful deep learning models  but without the need of collecting illuminant information, and relying instead on widely available annotations from other tasks. 
In experiments our method was, as expected, outperformed by learned methods which use ground truth data. However, surprisingly, in cross-dataset experiments our method obtains similar results as other learned based methods. As future research we plan to train Illuminant Estimation / Object Recognition networks for more complex illumination scenarios, e.g. with multiple illuminants. These scenarios are currently hardly investigated with learned based methods because of the lack of large datasets with ground truth information. However, in the proposed end-to-end learning of illuminant estimation the presence of ground truth data would not be required.

\minisection{Acknowledgements:} This work was supported by TIN2016-79717-R of the Spanish Ministry and the CERCA Programme / Generalitat de Catalunya.
We gratefully acknowledge the support of the NVIDIA Corporation with the donation of the Titan X Pascal GPU used for this research.

\bibliographystyle{IEEEbib}
\bibliography{cc000}

\end{document}